\documentclass[runningheads]{llncs}
\usepackage[T1]{fontenc}
\usepackage{graphicx}
\usepackage{booktabs}
\usepackage[misc]{ifsym}

\usepackage{subcaption} %
\usepackage{hyperref}       
\usepackage{url}            
\usepackage{amsfonts}       
\usepackage{nicefrac}       
\usepackage{microtype}      
\usepackage{xcolor}         
\usepackage{amsmath}
\usepackage{algorithm}
\usepackage{algorithmic}
\usepackage{paralist}

\usepackage{amsmath,amsfonts,bm}

\def\vtheta{{\bm{\theta}}}

\def\vpitheta{{\bm{\pi_{\theta}}}}

\def\vpithetaz{{\bm{\pi}_{\vtheta_0}}}

\def\eqref#1{equation~\ref{#1}}

\def\1{\bm{1}}

\def\vtheta{{\bm{\theta}}}

\def\vc{{\bm{c}}}

\def\vs{{\bm{s}}}

\def\vw{{\bm{w}}}
\def\vx{{\bm{x}}}
\def\vy{{\bm{y}}}

\DeclareMathAlphabet{\mathsfit}{\encodingdefault}{\sfdefault}{m}{sl}
\SetMathAlphabet{\mathsfit}{bold}{\encodingdefault}{\sfdefault}{bx}{n}

\usepackage{bm}
\usepackage{multirow}
\usepackage[inline, shortlabels]{enumitem}
\usepackage{caption}
\usepackage{bbding}

\usepackage[listings]{tcolorbox}
\tcbuselibrary{listings,theorems}
\definecolor{bluex}{rgb}{0.27, 0.42, 0.81}
\definecolor{purplex}{HTML}{9564bf}
\definecolor{red3}{HTML}{C52A20}
\definecolor{red2}{HTML}{B36A6F}
\definecolor{red1}{HTML}{FFb5b5}
\definecolor{purple}{HTML}{B36A6F}
\definecolor{darkyellow}{HTML}{D5BA82}
\definecolor{blue1}{HTML}{508AB2}
\definecolor{blue2}{HTML}{C4E4E3}
\definecolor{green1}{HTML}{A1D0C7}
\definecolor{green2}{HTML}{BFF6BA}
\definecolor{green3}{HTML}{028100}
\definecolor{teal}{HTML}{508AB2}
\definecolor{Gray}{gray}{0.94}
\definecolor{orange3}{HTML}{c28c69}
\definecolor{blue3}{HTML}{3b75af}

\definecolor{blue3}{HTML}{3B75AF}     
\definecolor{green3}{HTML}{4CAF50}    
\definecolor{red3}{HTML}{E74C3C}      
\definecolor{purple3}{HTML}{9B59B6}   
\definecolor{yellow3}{HTML}{F1C40F}   
\definecolor{cyan3}{HTML}{1ABC9C}     
\definecolor{pink3}{HTML}{FF69B4}     
\definecolor{brown3}{HTML}{8B4513}    
\definecolor{gray3}{HTML}{7F8C8D}     
\definecolor{darkblue}{HTML}{2C3E50}  
\definecolor{lightblue}{HTML}{5DADE2} 

\newtcolorbox{mybox}{colback=white!5!white,colframe=black!75!black, left=.05in, right=.05in}
\newtcbtheorem[]{exmp}{Example}%
{colback=green2!5,colframe=gray3,fonttitle=\bfseries, left=.02in, right=.02in,bottom=.02in, top=.02in}{exmp}
\newtcbtheorem{emptybox}{}%
{colback=green2!5,colframe=blue1,fonttitle=\bfseries,theorem style=plain, left=.0in, right=.0in,bottom=.02in, top=.02in,terminator sign none}{emptybox}

\def\vtheta{{\bm{\theta}}}

\def\vw{{\mathbf{w}}}

\def\vx{{\mathbf{x}}}
\def\vy{{\mathbf{y}}}

\def\vpiref{{\bm{\pi}_{\text{ref}}}}

\def\vpitheta{{\bm{\pi_{\theta}}}}

\def\vpithetandone{{\bm{\pi_{\theta_{n-1}}}}}

\def\vpithetaz{{\bm{\pi}_{\vtheta_0}}}

\newcommand{\mymethod}{MI-EPO}
\newcommand{\fullnamemymethod}{Multi-Objective Exploration and Preference Optimization via Mutual Information}

\begin{document}

\title{Multi-Objective Exploration and Preference Optimization via Mutual Information}

\titlerunning{MI-EPO}

\author{
Hongyan Xie\inst{1,2}\thanks{Work completed during an internship at Xingchen AGI Lab.} \and
Yikun Ban\inst{1} \and
Ruiyu Fang\inst{2} \and
Zixuang Huang\inst{1} \and
Deqing Wang\inst{1}\Letter\and
Jianxin Li\inst{1} \and
Shuangyong Song\inst{2}
}

\authorrunning{H. Xie et al.}

\institute{
School of Computer, Beihang University, Beijing, China
\email{\{xiehongyan,yikunb,huang\_zx,dqwang\}@buaa.edu.cn}
\and
Xingchen AGI Lab, China Telecom Artificial Intelligence Technology (Beijing) Co., Ltd, Beijing, China
\newline
\email{\{fangry,songshy\}@chinatelecom.cn}
}

\toctitle{MI-EPO: Mutual Information Enhanced Preference Optimization for Multi-Objective Alignment}

\tocauthor{Hongyan Xie, Yikun Ban, Ruiyu Fang, Zixuang Huang, Deqing Wang, Jianxin Li, Shuangyong Song}

\maketitle            
\begin{abstract}
Aligning large language models with diverse and heterogeneous human values requires multi-objective alignment methods to effectively trade off conflicting preference dimensions. Current methods achieve this trade-off by training policies conditioned on preference vectors and leveraging online direct preference optimization. However, exploration uncertainty can cause the reward distributions of responses generated under different preference vectors to overlap, and the generated responses may fail to effectively align with the corresponding preference vectors.
In this paper, we propose  Multi-Objective Exploration and Preference Optimization via Mutual Information (MI-EPO), an information-theoretic framework. It unifies multi-objective exploration and alignment by maximizing the joint conditional mutual information among generated responses, preference feedback, and preference vectors. By incorporating a probabilistic routing mechanism, MI-EPO  naturally decomposes objective alignment and preference-aware exploration, encouraging the model to generate responses that are distinguishable and aligned with different preference conditions.
Experiments on safe alignment and helpful assistant tasks show that MI-EPO significantly improves the alignment between generated responses and preference vectors, makes the outputs more controllable, and achieves stable trade-offs across multiple objectives. 

\keywords{Large Language Models  \and Multi-Objective Alignment \and Online Direct Preference Optimization \and Mutual Information}
\end{abstract}


\section{Introduction}

Reinforcement Learning from Human Feedback (RLHF) \cite{ouyang2022training} successfully aligns large language models (LLMs)~\cite{li2024tele,he2024telechat,wang2024telechat,wang2025technical,li202452b,liu2025training} with human values using Proximal Policy Optimization (PPO)~\cite{schulman2017proximal}.
However, most existing methods assume homogenized human values~\cite{bakker2022fine,huang2026real}, compressing diverse values into a single scalar reward. In reality,  human values are heterogeneous and multi-dimensional, and often require trade-offs across multiple potentially conflicting preference dimensions, such as helpfulness and harmlessness.

Recent research in language model alignment has increasingly moved from single-objective alignment to Multi-Objective Alignment (MOA). Some early methods extend RLHF to Multi-Objective RLHF (MORLHF)~\cite{rame2023rewarded,wang2024arithmetic,li2020deep}, where human feedback is decomposed into multiple preference dimensions and separate proxy reward models are trained for each objective. The language model policy is then optimized using PPO~\cite{schulman2017proximal}, with reward weights adjusted to trade off different preference objectives. However, PPO training suffers from instability and high computational cost~\cite{kumar2020discor,rafailov2023direct}. Motivated by the superior efficiency and stability of Direct Preference Optimization (DPO)~\cite{rafailov2023direct}, recent studies~\cite{zhou2024beyond,li2025self} have explored multi-objective alignment within the DPO framework. However, these methods still require training and maintaining multiple separate models.

\begin{figure}[t]
        \centering
    \begin{subfigure}[t]{0.34\columnwidth}
        \centering
        \includegraphics[width=\linewidth]{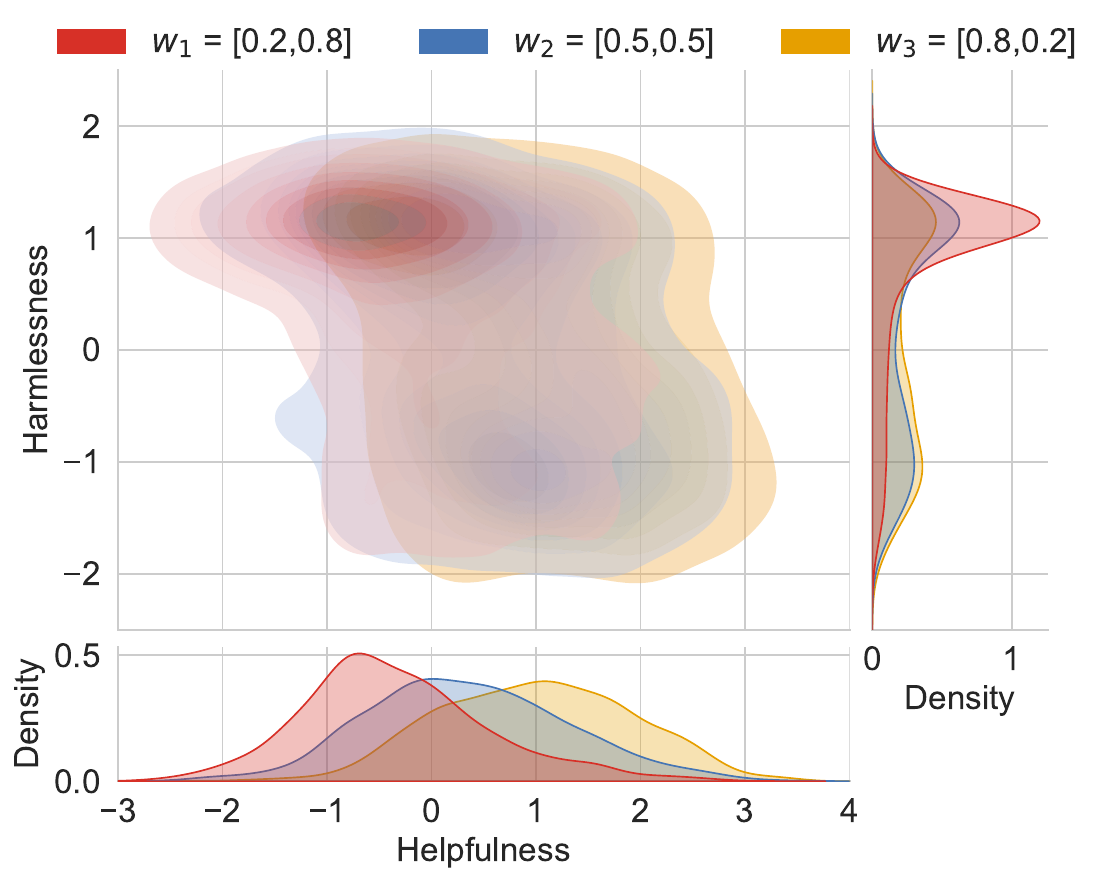}
        \caption{Reward distribution}
        \label{fig:mo_kde}
    \end{subfigure}
    \hfill
    \begin{subfigure}[t]{0.64\columnwidth}
        \centering
        \includegraphics[width=\linewidth]{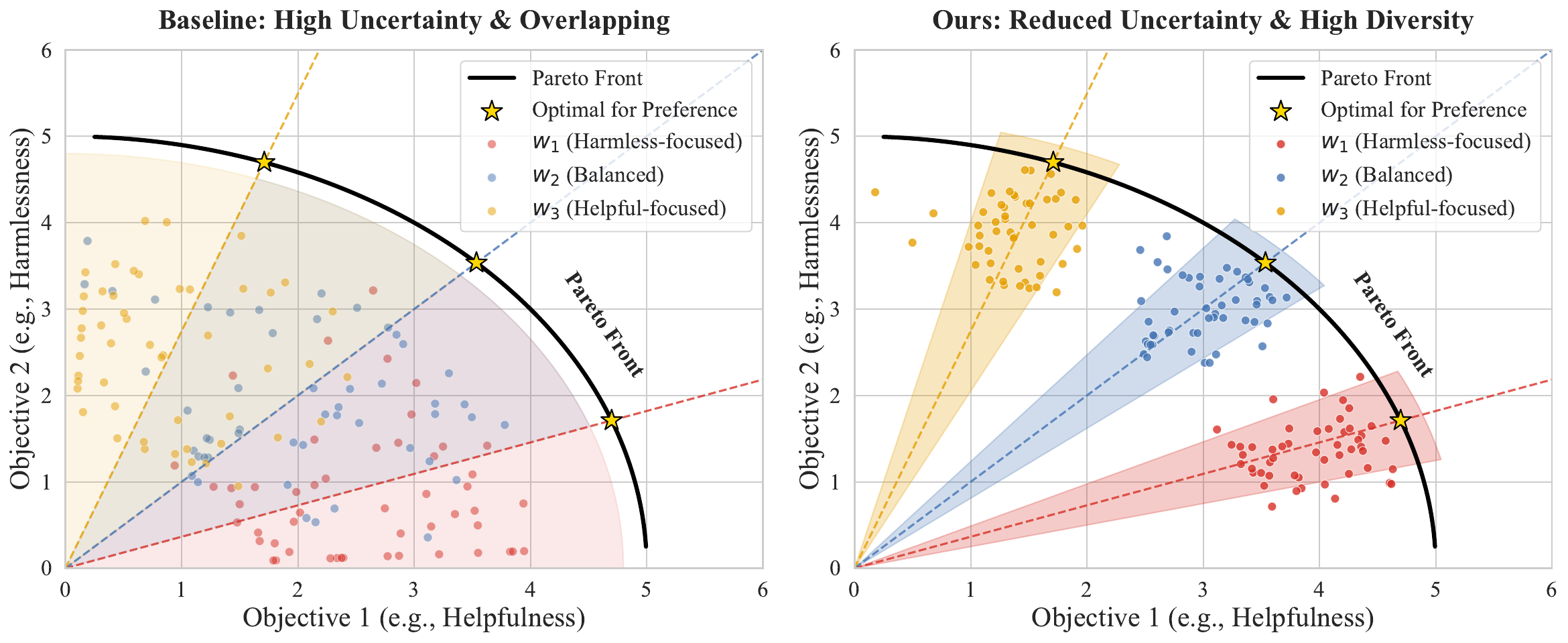}
        \caption{Exploration Uncertainty}
        \label{fig:mo_method}
    \end{subfigure}
    \caption{(a) Reward distributions of responses generated under different preference vectors during the test phase by the baseline method MO-ODPO. (b) The baseline method MO-ODPO  exhibits high exploration uncertainty during training, causing the exploration regions under different preference conditions to heavily overlap. In contrast, our method reduces exploration uncertainty, making the exploration regions corresponding to different preference conditions more distinguishable while maintaining overall diversity.}
  
    \label{fig:motivation}
\end{figure}

Recently, several works~\cite{yangrewards,yang2024metaaligner} leverage the instruction-following capability of LLMs by conditioning the policy on preference vectors or expected reward values. While such offline methods significantly reduce training cost, they rely on static preference datasets, which introduces distribution shift between the policy that generated the data and the policy being optimized~\cite{xu2024dpo,xiong2024iterative}. To address this limitation, recent work extends Online AI Feedback (OAIF) to the multi-objective setting and proposes Multi-Objective Online Direct Preference Optimization (MO-ODPO)~\cite{gupta2025robust}. 
By conditioning the policy on preference vectors, MO-ODPO is able to generate responses according to different preference vectors and combines the optimization stability of DPO with the exploration capability of online feedback, allowing effective trade-offs across multiple objectives. 
However, we found that even the currently best-performing MO-ODPO method exhibits substantial reward fluctuations in responses generated under the same preference conditions. Moreover, the reward distributions of responses under different preference conditions show significant overlap, as illustrated in Figure~\ref{fig:mo_kde}. This phenomenon suggests that the inherent exploration uncertainty during online generation weakens the conditional control of the preference vectors $W$ over the generated responses $Y$, resulting in a pronounced misalignment between them.

In this paper, we propose \fullnamemymethod\ (\mymethod), which unifies exploration and multi-objective alignment from an information-theoretic perspective by maximizing the joint conditional mutual information $\mathcal{I}(Y; C_Z, W, Z \mid X)$ to train a policy that generates responses conditioned on given preference vectors. Specifically, we formulate multi-objective alignment as maximizing the joint conditional mutual information $\mathcal{I}(Y; C_Z, W, Z \mid X)$, where $C_Z$ is the feedback of the objective selected through a probabilistic routing variable $Z$. By the chain rule of mutual information, this objective naturally decomposes into two complementary terms: $\mathcal{I}(Y; C_Z \mid X, W, Z)$ and $\mathcal{I}(Y; W \mid X)$. The former promotes objective-specific preference alignment, while the latter encourages preference-aware exploration, enabling the policy to generate distinguishable responses under different preference conditions (Figure~\ref{fig:mo_method}). 
The source code is publicly available at \url{https://github.com/jyxhyan/MI-EPO}.

Our contributions are summarized as follows:
\begin{itemize}
    \item We unify multi-objective alignment and exploration by maximizing the joint mutual information $\mathcal{I}(Y; C_Z, W, Z \mid X)$.
    \item We introduce a mutual information objective between responses and preference vectors, mitigating homogenized outputs under varying preferences and misalignment between generated responses and preference vectors.
    \item We demonstrate the effectiveness of our method through both theoretical analysis and empirical experiments.
\end{itemize}

\section{Preliminary}

\paragraph{DPO.} DPO is an offline preference optimization paradigm that circumvents the need for explicit reward modeling and reinforcement learning. By exploiting the closed-form optimal policy of a KL-constrained RL objective, DPO reparameterizes the reward function $r(\vx, \vy)$ as:\begin{equation}r(\vx, \vy) = \beta \log \frac{\vpitheta(\vy|\vx)}{\vpiref(\vy|\vx)} + \beta \log Z(\vx),\end{equation}where $\vpitheta$ is the parameterized policy, $\vpiref$ is the reference policy, and $Z(\vx)$ is the partition function.Given a preference dataset $\mathcal{D} = \{(\vx_i, \vy^+_i, \vy^-_i)\}_{i=1}^N$, where $\vy^+$ is preferred over $\vy^-$, DPO models pairwise preferences using the Bradley–Terry (BT) model. The preference probability is expressed as $P(\vy^+ \succ \vy^- \mid \vx) = \sigma(r(\vx, \vy^+) - r(\vx, \vy^-))$. Since the partition function $Z(\vx)$ cancels out in the reward difference, the DPO objective can be simplified to a binary cross-entropy loss:
\begin{equation}\label{eq:dpo_loss}\mathcal{L}_{\text{DPO}}(\vpitheta; \vx, \vy^+,\vy^-) = - \log \sigma \left( \beta \log \frac{\vpitheta(\vy^+|\vx)}{\vpiref(\vy^+|\vx)} - \beta \log \frac{\vpitheta(\vy^-|\vx)}{\vpiref(\vy^-|\vx)} \right).\end{equation}
This formulation allows for direct policy optimization using standard supervised learning techniques, improving training stability and computational efficiency.

\paragraph{Mutual Information Maximization.}

Let $X$ denote the input prompt, $Y$ the generated response, and $C \in \{0, 1\}$ a binary variable indicating human preference, where $c = 1$ denotes a preferred response and $c = 0$ a dispreferred one. The Conditional Mutual Information (CMI) \cite{ma2021conditional} quantifies the expected shared information between $Y$ and $C$ given $X$. Conditioning on $X$ isolates the relationship between the generation and its preference score by treating the prompt as a known constant. As direct computation of CMI is generally intractable, it is bounded from below by the InfoNCE objective \cite{tsai2022conditional}:
\begin{align}
    I(Y; C|X) 
    &:= \mathbb{E}_{\vx \sim X} \left[ D_{\text{KL}} \left( P_{Y,C|X=\vx} \parallel P_{Y|X=\vx} P_{C|X=\vx} \right) \right] \notag \\
    &\geq \text{InfoNCE} := \sup_{f} \mathbb{E} \left[ \log \frac{\exp(f(\vy, \vc))}{\exp(f(\vy, \vc)) + \sum_{j=1}^m \exp(f(\vy_j, \vc_j))} \right],
\end{align}
where Positive pairs $(y, c) \sim P_{Y,C|X}$, negative pairs $(y_j, c_j) \sim P_{Y|X} P_{C|X}$, and $f$ denotes a score function.  Given a prompt distribution $p(x)$ and a policy $\pi_\theta$, a prompt $x \sim p(x)$, a positive pair $(y^+, c) \sim \pi_\theta(y, c=1 \mid x)$, and a negative pair $(y^-, c) \sim \pi_{\text{ref}}(y \mid x)\, p(c \mid x)$ are sampled.
Utilizing a parameterized critic function $f_\phi$, the InfoNCE objective with preference feedback simplifies to a pairwise contrastive loss:
\begin{align}
    \mathcal{L}_{\text{InfoNCE}}(\phi; \vy^+, \vy^-) = -\log \frac{\exp(f_\phi(\vy^+, \vc=1))}{\exp(f_\phi(\vy^+, \vc=1)) + \exp(f_\phi(\vy^-, \vc=0))}.
\end{align}
When $f_\phi(\vx,\vy)=\beta \log \frac{\pi_\theta(\vy \mid \vx)}{\pi_{\text{ref}}(\vy \mid \vx)}$, the InfoNCE objective reduces to Equation~\ref{eq:dpo_loss}.

\section{Method}
In this section, we introduce the \fullnamemymethod.
\subsection{Motivations and Problem Formulation}
\label{sec:problem}
We train a prompt-conditioned controllable generation policy $\vpitheta(\vy \mid \vx, \vw)$, where the preference vector $\vw \in \mathbb{R}^K$ is incorporated as part of the policy input. Each objective $k$ is associated with a reward model $R^k(\vx,\vy)$ that produces an objective-specific scalar reward $s^k$. The preference vector $\vw = [w_1, \ldots, w_K]^\top$ lies on the probability simplex
$\Delta^{K-1} = \{\, w \mid \sum_{k=1}^K w_k = 1,\; w_k \ge 0 \,\}$,
where $w_k$ represents the relative importance assigned to the $k$-th objective during generation.

Online multi-objective alignment is typically formulated as a sequential procedure. A single prompt $\vx$ is sampled from a dataset $\mathcal{D}$, and a preference vector $\mathbf{w}$ is sampled from the simplex $\Delta^{K-1}$ to condition the prompt. Two candidate responses, $\vy^1$ and $\vy^2$, are generated from the current policy $\pi_\theta(\cdot \mid \vx, \mathbf{w})$. For each objective $k$, the candidate response with the higher reward score is designated as the preferred response $\vy^{+,k}$, and the other as the non-preferred $\vy^{-,k}$. We define an objective-specific preference tuple $(\vx, \vy^{+,k}, \vy^{-,k})$ for each objective $k$.
To find a Pareto optimal solution, the multi-objective optimization (MOO) is scalarized into a weighted sum:
\begin{align}
    \min_{\theta} \sum_{k=1}^{K} w_{k} \, \mathcal{L}_\text{DPO}\!\left(\vpitheta; (\vx, \vw), \vy^{+,k}, \vy^{-,k} \right)
\quad
\text{s.t.} \quad \sum_{k=1}^K w_{k} = 1,\; w_{k} \ge 0.
\end{align}
where $w_{k}$ denotes the $k$-th component of the sampled vector $\vw$.

\begin{definition}
    A solution $\theta^*$ is Pareto optimal if there exists no other solution that is at least as good in all objectives and strictly better in at least one.
\end{definition}

However, the online exploration process based on policy sampling faces significant challenges. Although the Pareto frontier defines the ideal trade-off boundary among objectives, the inherent exploration uncertainty during online generation weakens the conditional control of the preference Vectors $W$ over the generated responses $Y$, leading to severe misalignment between them. This issue manifests in two aspects: (1) the generated responses may deviate from the intended preference vectors; and (2) responses generated under different preference conditions may substantially overlap in the reward distributions.

To address these challenges, we unify multi-objective exploration and alignment from an information-theoretic perspective. Let $X$, $Y$, and $W$ denote the prompt, the policy-sampled response, and the conditional preference vector, respectively. For each objective $k$, let $C_k \in \{0,1\}$ denote its preference feedback signal. As these signals are typically produced independently by different reward models, we follow the standard multi-objective evaluation setting and assume that the feedback variables $\{C_k\}_{k=1}^K$ are conditionally independent given $(X,Y)$. To model how preference vectors select among objectives, we adopt a probabilistic objective routing formulation. We introduce an auxiliary variable $Z \in \{1,\ldots,K\}$ determined by the preference vector such that $P(Z = k \mid W) = w_k$. The variable $Z$ indicates the objective emphasized by the current response. Let $C_Z$ denote the feedback associated with the routed objective. The overall alignment objective can then be expressed as maximizing the joint conditional mutual information $\mathcal{I}(Y; C_Z, W, Z \mid X)$. Applying the chain rule of mutual information, and recognizing that $Z$ is conditionally independent of $Y$ given $W$ (i.e., $\mathcal{I}(Y; Z \mid X, W) = 0$), we can exactly decompose the objective:
\begin{align}
\label{eq:mi}
J(\theta) 
&= \mathcal{I}(Y; C_Z, W, Z \mid X) \notag \\
&= \mathcal{I}(Y; W \mid X) 
   + \mathcal{I}(Y; Z \mid X, W) 
   + \mathcal{I}(Y; C_Z \mid X, W, Z) \notag \\
&= \mathcal{I}(Y; W \mid X) 
   + \mathbb{E}_{z \sim P(Z|W)} 
   \left[ \mathcal{I}(Y; C_z \mid X, W, Z=z) \right] \notag \\
&= \mathcal{I}(Y; W \mid X) 
   + \sum_{k=1}^K w_k \, \mathcal{I}(Y; C_k \mid X, W), \\
\max_{\theta}\; & J(\theta) \quad 
\text{s.t.} \quad \sum_{k=1}^K w_k = 1,\; w_k \ge 0.
\end{align}
where maximizing the conditional mutual information $\mathcal{I}(Y; C_k \mid X, W)$ enables objective-specific preference alignment, while maximizing $\mathcal{I}(Y; W \mid X)$ reduces exploration uncertainty and encourages responses aligned with preference vectors, yielding distinguishable and diverse outputs across preferences.

\paragraph{Information-theoretic interpretation.}
We next provide an information-theoretic interpretation of the proposed objective, highlighting its implications for preference identifiability and structured exploration.

The term $\mathcal{I}(Y; W \mid X)$ measures how much information the generated response $Y$ preserves about the preference vector $W$. Using the identity
$\mathcal{I}(Y; W \mid X) = H(W \mid X) - H(W \mid Y, X)$,
we see that, under a fixed prior entropy $H(W \mid X)$, maximizing this term reduces the posterior uncertainty $H(W \mid Y, X)$. This implies that the generation process becomes more informative with respect to preference conditions, improving identifiability and reducing ambiguity across different preference-induced behaviors.

To understand the effect on exploration, consider the marginal entropy of the policy distribution $H(Y \mid X)$ induced by the preference prior $P(W)$. Using the standard mutual information decomposition,
$H(Y \mid X) = \mathcal{I}(Y; W \mid X) + H(Y \mid W, X)$,
we observe that $\mathcal{I}(Y; W \mid X)$ provides a structural lower bound on the marginal entropy up to the conditional entropy term. Therefore, increasing $\mathcal{I}(Y; W \mid X)$ encourages higher diversity in $Y$ while maintaining consistency conditioned on $W$. This results in a partitioned exploration space where different preference vectors induce distinguishable yet diverse generation modes.

Together, these effects motivate our design, which jointly promotes preference alignment and structured exploration within a unified information-theoretic framework.

\subsection{Maximizing $\mathcal{I}(Y; C_k \mid X, W)$ for Preference Alignment}

Prior work \cite{xiao2025infopo} demonstrated that under the Bradley–Terry (BT) model assumption, the objective of DPO can be interpreted as maximizing the conditional mutual information between the response $Y$ and the preference feedback $C_k$, where InfoNCE~\cite{oord2018representation} serves as the variational estimator.  Therefore, for each objective $k$, we adopt DPO to maximize $\mathcal{I}(Y; C_k \mid X, W)$, which encourages the policy to generate responses that align with the objective-specific preference feedback. The resulting objective can be formulated as follows:
\begin{equation}
\label{eq:yc_loss}
\mathcal{L}_{\text{YC}}(\vpitheta; \vy^{+,k}, \vy^{-,k}) \! = \! - \! \log \sigma \! \left( \beta_c \log \frac{\vpitheta(\vy^{+,k}|\vx, \vw^+)}{\vpiref(\vy^{+,k}|\vx, \vw^+)} \!- \!\beta_c \log \frac{\vpitheta(\vy^{-,k}|\vx, \vw^+)}{\vpiref(\vy^{-,k}|\vx, \vw^+)} \!\right),
\end{equation}
where $\mathbf{w}^+$ denotes the preference vector used for sampling $\vy^{+,k}$ and $\vy^{-,k}$.

\subsection{Maximizing $\mathcal{I}(Y; W \mid X)$ for Reducing Exploration Uncertainty}
To maximize the CMI $\mathcal{I}(Y; W \mid X)$, we derive a tractable lower bound based on the InfoNCE framework:
\begin{align}
    \mathcal{I}(Y; W \mid X) &:= \mathbb{E}_{x \sim X} \left[ D_{\text{KL}} \left( P_{Y,W \mid X=\vx} \parallel P_{Y \mid X=\vx} P_{W \mid X=\vx} \right) \right] \notag \\
    & \geq \sup_{g} \mathbb{E} \left[ \log \frac{\exp(g( \vy, \vw))}{\exp(g( \vy, \vw)) + \sum_{j=1}^m \exp(g(\vy, \vw_j))} \right],
\end{align}
where $g$ represents a score function. In this formulation, the positive pair $(\vy, \vw) \sim P_{Y,W \mid X=\vx}$ is drawn from the conditional joint distribution, indicating that the response $\vy$ satisfies the specified preference condition $\vw$. Conversely, the $m$ negative pairs $(\vy, \vw_j) \sim P_{Y \mid X=\vx}P_{W \mid X=\vx}$, representing cases where the response $\vy$ is not aligned with the preference vector $\vw_j$. 
Given a prompt distribution $p(\vx)$ and a policy $\vpitheta$, a prompt $\vx \sim p(\vx)$, a positive pair $(\vy, \vw^+) \sim \vpitheta(\vy \mid \vx, \vw^+)$ are sampled. To form the negative pair, we reuse the same response \(\vy\) and pair it with an independently drawn condition \(\vw^- \sim p(\vw \mid \vx)\). The InfoNCE  loss under the conditional preference vector is defined as follows:
\begin{align}
    \mathcal{L}_{\text{YW}}(\phi; \vw^+,\vw^-) = -\log \frac{\exp(g_\phi(\vy, \vw^+))}{\exp(g_\phi(\vy, \vw^+)) + \exp(g_\phi(\vy, \vw^-))}.
\end{align}

To instantiate the parametric critic function $g_\phi$, the implicit reward parameterization is adopted \cite{rafailov2023direct}. Specifically, $g_\phi(\vy, \vw)$ is defined as follows:
\begin{align}
    \label{eq:gphi}
    g_\phi(\vy, \vw) = \beta_w \log \frac{\vpitheta(\vy | \vx, \vw)}{\pi_{\text{ref}}(\vy | \vx, \vw)},
\end{align}
Substituting this parameterization into Equation~\ref{eq:gphi} yields the following InfoNCE loss function:
\begin{align}
    \label{eq:yw_loss}
    \mathcal{L}_{\text{YW}}(\vpitheta; \vw^+,\vw^-) = -\log \sigma \left( \beta_w \log \frac{\vpitheta(\vy | \vx, \vw^+)}{\vpiref(\vy | \vx, \vw^+)} - \beta_w \log \frac{\vpitheta(\vy | \vx, \vw^-)}{\vpiref(\vy | \vx, \vw^-)} \right),
\end{align}

\subsection{\fullnamemymethod}

Based on the joint mutual information decomposition formula proposed in Section~\ref{sec:problem}, we scalarize and integrate the preference alignment loss $\mathcal{L}_{YC}$ with the exploration enhancement loss $\mathcal{L}_{YW}$. The final MI-EPO loss function is defined as follows:
\begin{align}
\label{eq:final_loss}
\mathcal{L}_\text{\mymethod} &= \sum_{k=1}^K w_k \mathcal{L}_{\text{YC}}(\vpitheta; \vy^{+,k}, \vy^{-,k})  + \frac{1}{2} \sum_{y \in {y^1, y^2}} \mathcal{L}_{\text{YW}}(\vpitheta; \vw^+,\vw^-) \notag \\
&\!\!\!\!= - \sum_{k=1}^K w_k \log \sigma \left( \beta_c \log \frac{\vpitheta(\vy^{+,k}|\vx, \vw^+)}{\vpiref(\vy^{+,k}|\vx, \vw^+)} - \beta_c \log \frac{\vpitheta(\vy^{-,k}|\vx, \vw^+)}{\vpiref(\vy^{-,k}|\vx, \vw^+)} \right) \notag \\
&\!\!\!\!\quad - \frac{1}{2} \sum_{\vy \in {\vy^1, \vy^2}} \log \sigma \left( \beta_w \text{sg}\left(\log \frac{\vpitheta(\vy | \vx, \vw^+)}{\vpiref(\vy | \vx, \vw^+)}\right) - \beta_w \log \frac{\vpitheta(\vy | \vx, \vw^-)}{\vpiref(\vy | \vx, \vw^-)} \right) \notag \\
& \text{s.t.} \quad \sum_{k=1}^K w_{k} = 1,\; w_{k} \ge 0,
\end{align}
where the stop-gradient operator $\text{sg}(\cdot)$ ensures that the gradient of $\mathcal{L}_{\text{YW}}$ only flows through the counterfactual condition $\vw^-$, preventing overfitting to the self-generated response $\vy$. Since $\mathcal{L}_{\text{YW}}$ is independent of preference feedback, both $\vy^1$ and $\vy^2$ generated under $\vw^+$ can serve as valid anchors.

\subsection{Implementation}
Algorithm~\ref{alg:algorithm} presents the procedure of our proposed \mymethod, with detailed steps provided below: \begin{inparaenum}[(1)] \item We sample preference vectors $\vw^+$ and $\vw^-$  from a Dirichlet distribution defined over the probability simplex, where the concentration parameters $\boldsymbol{\alpha}$ control the sparsity and balance of the sampled weights. Subsequently, we define a prompt-construction function to formalize the integration of the user prompt $\vx$ with a preference vector $\mathbf{w}$:
\[G(\vx, \vw) =
\text{Human:}\{\vx\} \;\oplus\; \text{RN}_1 \{w_1\} \;\dots\; \text{RN}_K \{w_K\} \;\oplus\; \text{Assistant:},\]
where $\text{RN}_k$ denotes textual tokens representing the $k$-th preference dimension, and $\oplus$ indicates sequence concatenation.
\item  Given the current policy, we sample two responses $\vy^1$ and $\vy^2$. Each response is evaluated independently on $K$ objectives. We compare the two responses along each objective $k$ and label the response with the higher score as the preferred response $\vy^{+,k}$, while treating the other as the non-preferred response $\vy^{-,k}$.
\item  Subsequently, we compute the loss according to Equation~\ref{eq:final_loss} and update the policy parameters accordingly.
\end{inparaenum}

\begin{algorithm}[tb]
\caption{\mymethod\, for Online Multi-Objective Alignment}
\label{alg:algorithm}
\begin{algorithmic}[1]
\REQUIRE
initial policy $\vpithetaz$;
reward models $\{R^k\}_{k=1}^{K}$;
training dataset $\mathcal{D}$; Num training epochs $N$
    \FOR{$n := 1$ to $N$}
        \FOR{each prompt $\vx \in \mathcal{D}$}
            \STATE Sample preference vectors $\vw^+, \vw^- \sim \text{Dirichlet}(\alpha)$
            \STATE Sample two candidate responses $\vy^1, \vy^2 \sim \vpithetandone(\cdot \mid G(\vx, \vw^+))$
            \FOR{$k = 1$ \textbf{to} $K$}
                \STATE Compute the reward scores $s^{1,k}$ and $s^{2,k}$ for $(\mathbf{x}, \mathbf{y}^1)$ and $(\mathbf{x}, \mathbf{y}^2)$ using the reward model $R^k$
                \STATE The response with the higher score is denoted as $\mathbf{y}^{+,k}$, while the other is denoted as $\mathbf{y}^{-,k}$
                
            \ENDFOR
            \STATE Compute the loss via Equation \ref{eq:final_loss} and update the policy parameters via gradient descent
        \ENDFOR
    \ENDFOR

\end{algorithmic}
\end{algorithm}

\section{Experiments}
\label{sec:experiments}
We evaluate the performance of our \mymethod\ on two widely studied multi-objective alignment (MOA) tasks: safety alignment and helpful assistant alignment. These tasks involve multiple distinct and potentially competing objectives.

\subsection{Safety Alignment} \label{sec:safety}

\paragraph{Data and Training Setup.}
Multi-objective alignment aims to achieve effective trade-offs across multiple preference dimensions, enabling models to satisfy diverse user preferences. In this section, we focus on the safety alignment task, which involves two key alignment dimensions: \textit{helpfulness} and \textit{harmlessness}. To study this problem, we adopt the \texttt{PKU-SafeRLHF-10K} dataset~\cite{ji2023beavertails}, which provides preference annotations for both helpfulness and harmlessness on each question–answer pair. We randomly split the dataset into two subsets: $8$K samples for training and the remaining $2$K for testing.
In this experiment, we define two special tokens, $\mathrm{RN}_1$ and $\mathrm{RN}_2$, corresponding to the strings \texttt{helpfulness} and \texttt{harmlessness}, respectively.
For the backbone language model, we use Alpaca-7B. Following~\cite{lin2025parm}, we employ two open-source reward models as evaluation oracles to score each response along the helpfulness and harmlessness dimensions. The open-source reward models and datasets are publicly accessible, as detailed in Appendix A.  Additional details on training procedures and hyperparameter settings are reported in Appendix B.

\paragraph{Evaluation.} We evaluate performance by examining the Pareto fronts produced by different methods, specifically through the average normalized test reward curves. In addition, we employ three multi-objective metrics to evaluate the performance of different methods:
\begin{inparaenum}[(1)]
\item Hypervolume (HV)~\cite{zitzler1998multiobjective}, which measures the volume of the non-dominated region in the objective space; larger HV indicates better diversity and convergence;
\item Mean Inner Product (MIP)~\cite{lin2025parm}, which computes the average inner product between the preference vector and the reward vector, measuring how well generated solutions align with user preferences. Larger MIP indicates better alignment.
\item Conditional Reward Dispersion (CRD), which computes the average determinant of the covariance matrix of reward vectors under each preference condition, measuring the intra-condition dispersion of obtained rewards. Smaller CRD indicates more stable conditional control.
\end{inparaenum} Further details regarding these metrics and the evaluation procedure can be found in Appendix C.

\paragraph{Baselines.} The proposed \mymethod\ is compared with the following methods: \begin{inparaenum}[(1)]
\item Rewarded Soups (RS)~\cite{rame2023rewarded}, which fine-tunes $k$ base models and merges them into a single model in the parameter space according to the given preference vector $\vw$ during inference;
\item RiC~\cite{yangrewards}, which conditions the policy model on multiple contextual rewards and applies supervised fine-tuning for alignment;
\item MO-ODPO~\cite{gupta2025robust}, a multi-objective online DPO algorithm that aligns a single policy with diverse and potentially conflicting preferences via preference-conditioned prompts.
\end{inparaenum}
\begin{figure}[t]
    \centering
    \begin{subfigure}[b]{0.32\textwidth}
        \centering
        \includegraphics[width=\textwidth]{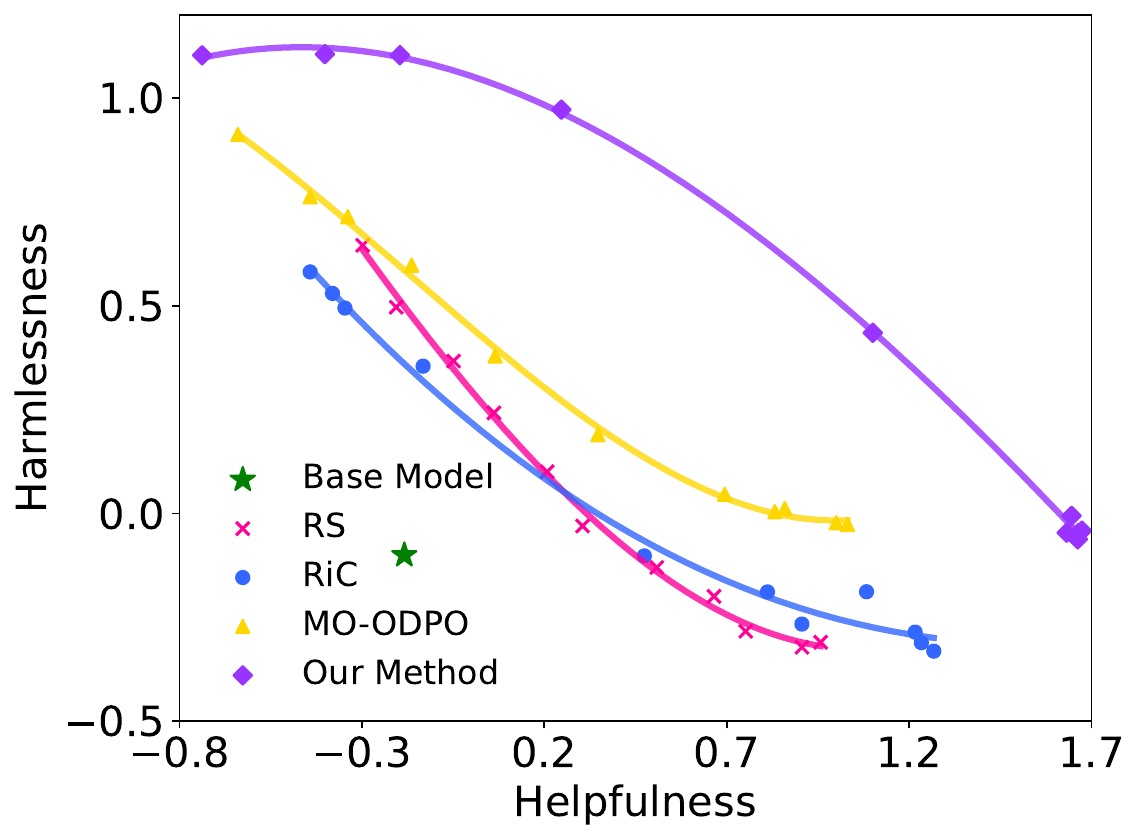}
        \caption{Pareto fronts}
        \label{fig:safe_pf}
    \end{subfigure}
    \hfill
    \begin{subfigure}[b]{0.64\textwidth}
        \centering
        \includegraphics[width=\textwidth]{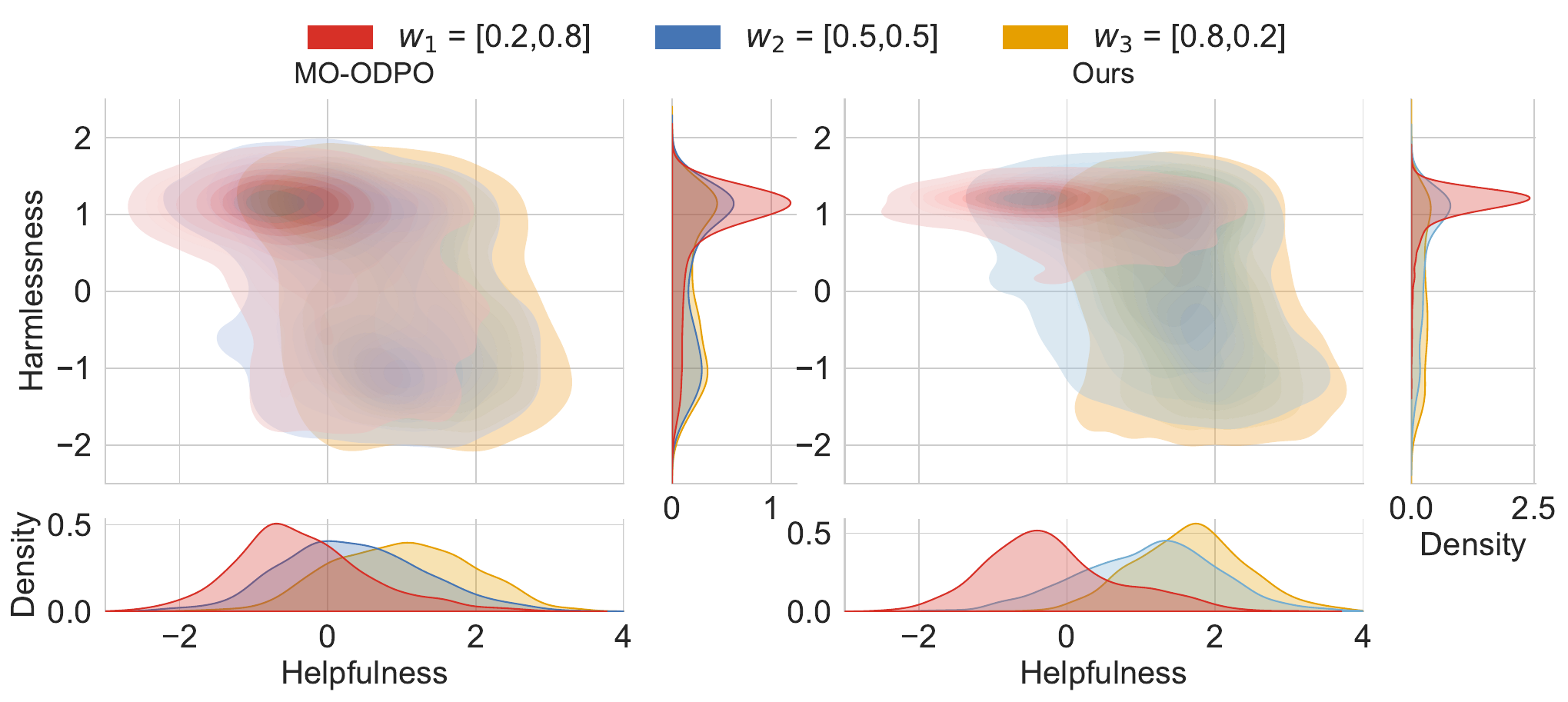}
        \caption{Reward distribution}
        \label{fig:safe_rd}
    \end{subfigure}

    \caption{(a) Pareto front curves produced by \mymethod\ and baseline methods on the safety alignment task. (b) Reward distributions of \mymethod\ and the baseline method MO-ODPO on the safety alignment task}
    \label{fig:safe_main}
\end{figure}

\begin{table}
\caption{Performance comparison on the Safety Alignment task using HV, MIP, and CRD metrics. Metrics with $\uparrow$ indicate that larger values are better, and metrics with $\downarrow$ indicate that smaller values are better.
}
\label{tab:safe_main}
    \begin{center}
                \setlength{\tabcolsep}{4pt}
                \begin{tabular}{lccc}
                \toprule
                & HV$\uparrow$ & MIP$\uparrow$ & CRD$\downarrow$\\
                \midrule
                RS \cite{rame2023rewarded} & 1.19 & 0.43 & 0.64 \\
                RiC \cite{yangrewards} & 1.13 & 0.58 & 0.89 \\
                MO-ODPO \cite{gupta2025robust} & 1.54 & 0.82 & 0.62\\
                \textbf{\mymethod}  & \textbf{2.70} & \textbf{1.01} & \textbf{0.29} \\
                \bottomrule
                \end{tabular}
    \end{center}
\end{table}

\paragraph{Results.}

Figure~\ref{fig:safe_pf} presents the empirical Pareto fronts achieved by \mymethod\ and the baseline methods. Each point represents the average multi-objective reward on the test set under a specific preference condition. Compared to the baselines, \mymethod\ produces Pareto fronts that cover a larger area. We compare the reward distributions of the baseline method MO-ODPO and our proposed method under different preference vectors $w$. As shown in Figure~\ref{fig:safe_rd}, for MO-ODPO, the reward distributions corresponding to different preference vectors exhibit substantial overlap. In contrast, the reward distributions generated by our method are more distinctly separated in the space. Notably, in the marginal distributions, when the preference vector emphasizes a specific dimension, the corresponding reward distribution shows a clear positive shift, providing strong evidence for the effectiveness of our method in multi-objective alignment tasks. Furthermore, our method substantially reduces the variance of reward distributions under each preference condition, effectively suppressing uncertainty in the generation process and enabling effective trade-offs among different objectives.

Table~\ref{tab:safe_main} reports quantitative evaluation results using  HV,  MIP, and CRD. \mymethod\ consistently outperforms all baselines across all metrics. Specifically, compared with the currently best-performing baseline method, MO-ODPO, \mymethod\ achieves a 68.8\% improvement in HV, indicating broader coverage and higher diversity of the solution set; a 23.2\% improvement in MIP, showing that the generated responses follow the specified preference vectors more closely; and a 53.2\% reduction in CRD indicates decreased variance of rewards under the same preference condition, resulting in more stable conditional control.  Moreover, when $\beta_w = 0$, \mymethod\ is equivalent to MO-ODPO. These results collectively confirm that \mymethod\ effectively mitigates cross-objective interference while achieving robust and balanced multi-objective trade-offs.

\subsection{Helpful Assistant}

\paragraph{Data and Training Setup.} 
In this section, we conduct an empirical evaluation of the helpful assistant task across three objectives: helpfulness, harmlessness, and humor. We use the \texttt{HH-RLHF} dataset~\cite{bai2022training} and adopt \texttt{Qwen3-8B} as the backbone language model. From approximately 160K multi-turn dialogues, 8K are randomly selected to form the training set, while 2K randomly sampled dialogues are used for testing during inference. 
In this experiment, we define three special tokens, $\mathrm{RN}_1$, $\mathrm{RN}_2$, and $\mathrm{RN}_3$, corresponding to the strings \texttt{helpfulness}, \texttt{harmlessness}, and \texttt{humor}, respectively.
Three independent, open-source reward models serve as evaluation oracles, providing an average score for each objective dimension. Detailed descriptions of the experimental setup and hyperparameters are provided in Appendices D and E.

\paragraph{Baselines and Evaluation.} We compare the proposed \mymethod\ with the same baseline methods used in the safety alignment task. The evaluation procedure can be found in Appendix C.

\begin{figure}[t]
    \centering
    \begin{subfigure}[b]{0.32\textwidth}
        \centering
        \includegraphics[width=\textwidth]{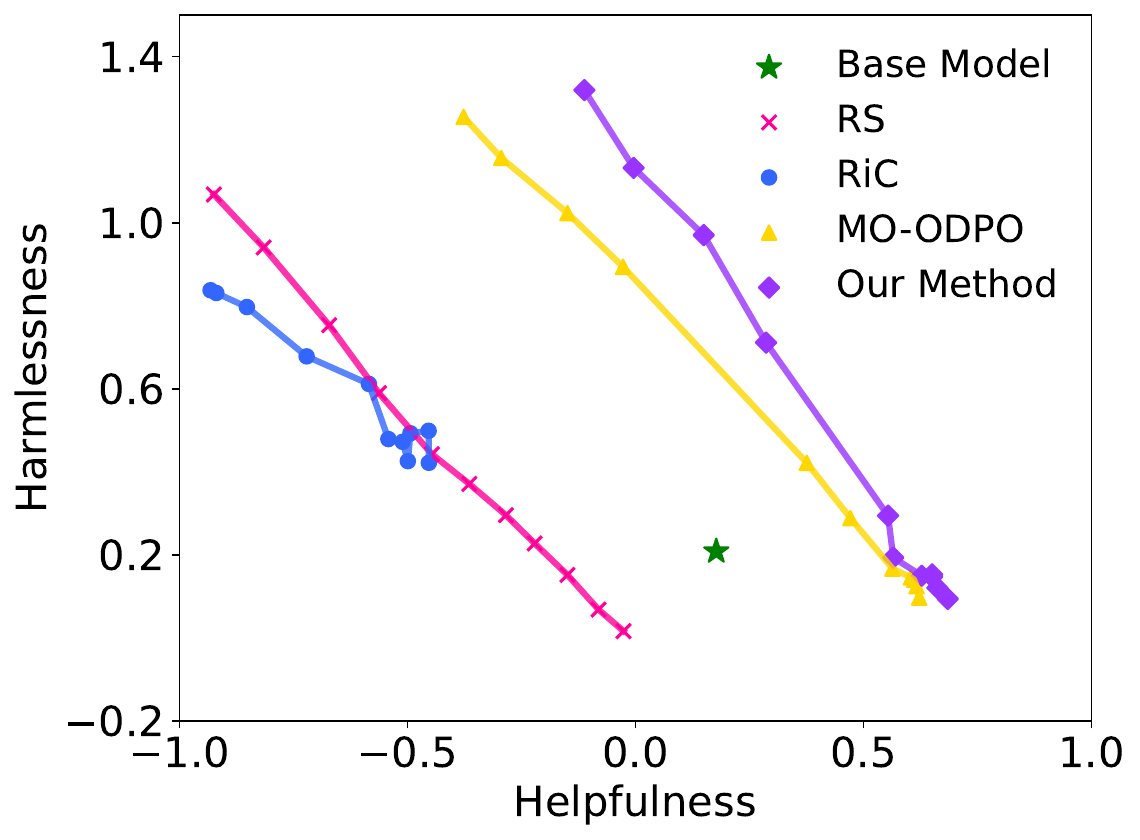}
        \caption{Pareto fronts}
        \label{fig:hh_pf}
    \end{subfigure}
    \hfill
    \begin{subfigure}[b]{0.64\textwidth}
        \centering
        \includegraphics[width=\textwidth]{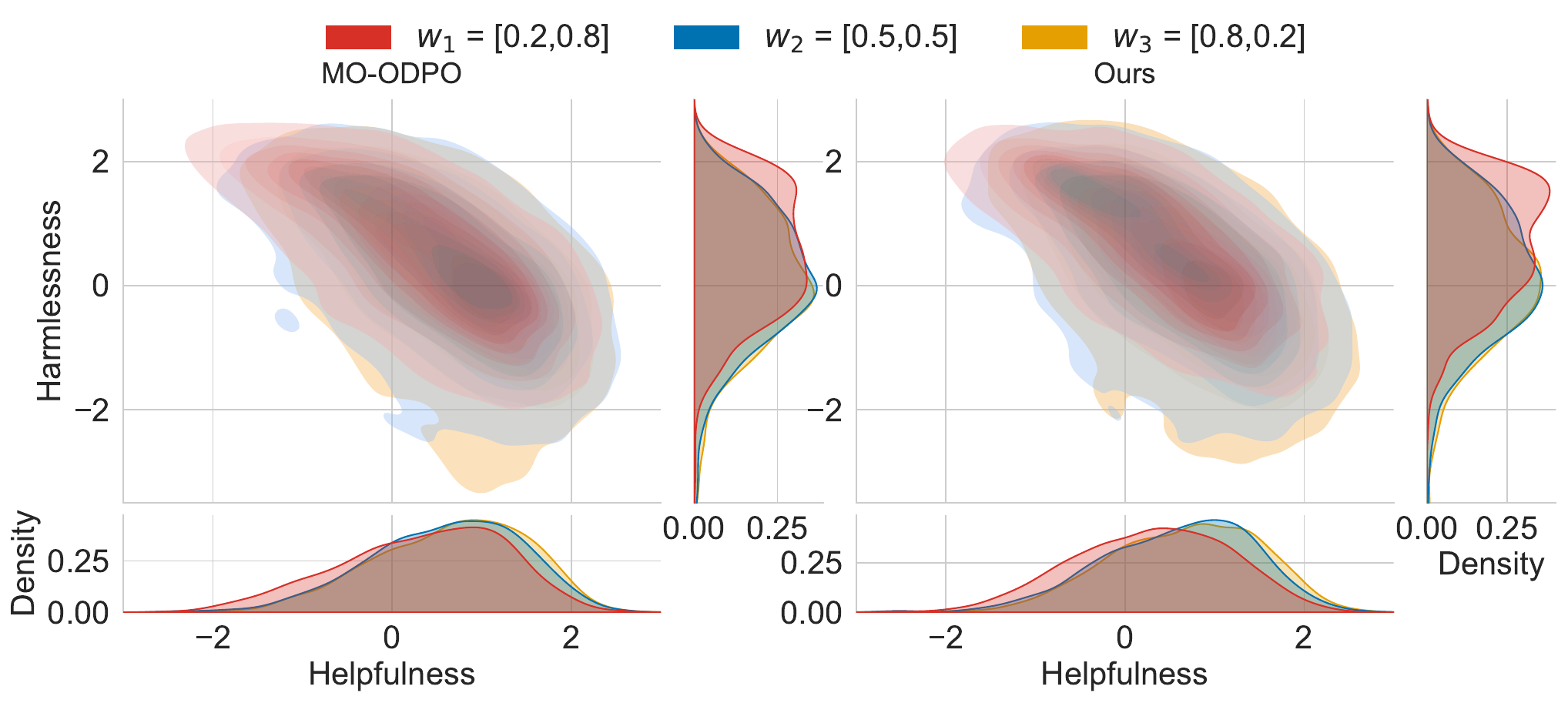}
        \caption{Reward distribution}
        \label{fig:hh_rd}
    \end{subfigure}

    \caption{(a) Pareto front curves produced by \mymethod\ and baseline methods on the helpful assistant task. (b) Reward distributions of \mymethod\ and the baseline MO-ODPO on the helpful assistant task. All methods are trained on the \texttt{Qwen3-8B-Base} model.}
    \label{fig:hh_main}
\end{figure}

\begin{table*}[ht]
 \caption{Performance comparison on the Helpful Assistant task across different objectives. Metrics with $\uparrow$ indicate that larger values are better, and metrics with $\downarrow$ indicate that smaller values are better.}

    \centering
    \begin{subtable}[t]{0.48\textwidth} 
        \centering
        \caption{Helpfulness and Harmlessness}
        \label{tab:harmless}
        \setlength{\tabcolsep}{4pt} 
        \begin{tabular}{lccc}
        \toprule
        & HV$\uparrow$  & MIP$\uparrow$  & CRD$\downarrow$ \\
        \midrule
        RS \cite{rame2023rewarded} & 0.68 & 0.21 & 0.58\\
        RiC \cite{yangrewards} & 0.46 & 0.07 & 1.05\\
        MO-ODPO \cite{gupta2025robust} & 1.68 & 0.64 & 0.57 \\
        \textbf{\mymethod} & \textbf{1.98} & \textbf{0.68} & \textbf{0.52} \\
        \bottomrule
        \end{tabular}
    \end{subtable}
    \begin{subtable}[t]{0.49\textwidth}
        \centering
        \caption{Helpfulness, Harmlessness, and Humor}
        \label{tab:hh-3d}
        \setlength{\tabcolsep}{4pt}
        \begin{tabular}{lccc}
        \toprule
        Method & HV$\uparrow$ & MIP$\uparrow$ & CRD$\downarrow$ \\
        \midrule
        RS \cite{rame2023rewarded} & 1.82 & 0.01 & \textbf{0.36}\\
        RiC \cite{yangrewards} & 2.00 & 0.15 & 0.63 \\
        MO-ODPO \cite{gupta2025robust} & 3.12 & 0.27 & 0.47 \\
        \textbf{\mymethod} & \textbf{5.84} & \textbf{0.35} & 0.44 \\
        \bottomrule
        \end{tabular}
    \end{subtable}
  
\end{table*}

\paragraph{Results.} 
We evaluate the ability of different methods to balance harmlessness and helpfulness in multi-turn dialogues. As shown in Figure~\ref{fig:hh_pf}, the Pareto front generated by \mymethod\ covers a broader region in the objective space. Furthermore, as illustrated in Figure~\ref{fig:hh_rd}, \mymethod\ produces reward distributions under different preference vectors that are more distinctly separated than those of the currently best-performing method, MO-ODPO, thereby enabling finer-grained preference control. In single-preference conditions, the variance of the reward distribution is also lower, which enhances the stability of conditional control and reduces uncertainty during the sampling process. These observations are consistent with the quantitative results reported in Table~\ref{tab:harmless}. Overall, these results indicate that \mymethod\ enables the policy model to more effectively trade off among multiple objectives.

\begin{figure*}[t]
    \centering
    \begin{subfigure}[b]{0.45\textwidth}
        \centering
        \includegraphics[width=\textwidth]{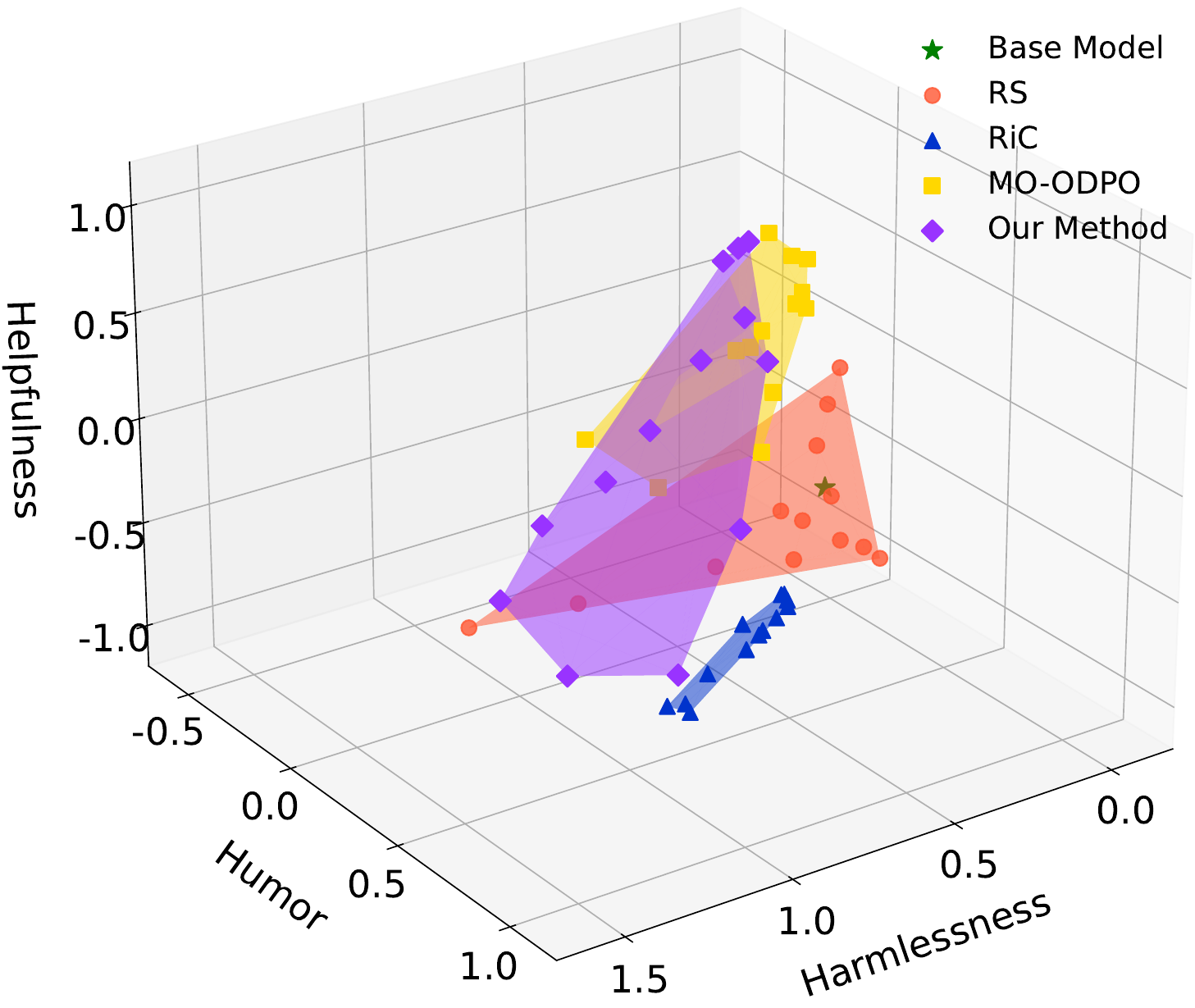}
        \caption{The learned Pareto frontier in three-dimensional space.}
        \label{fig:hh_3d}
    \end{subfigure}
    \hfill
    \begin{subfigure}[b]{0.45\textwidth}
        \centering
        \includegraphics[width=\textwidth]{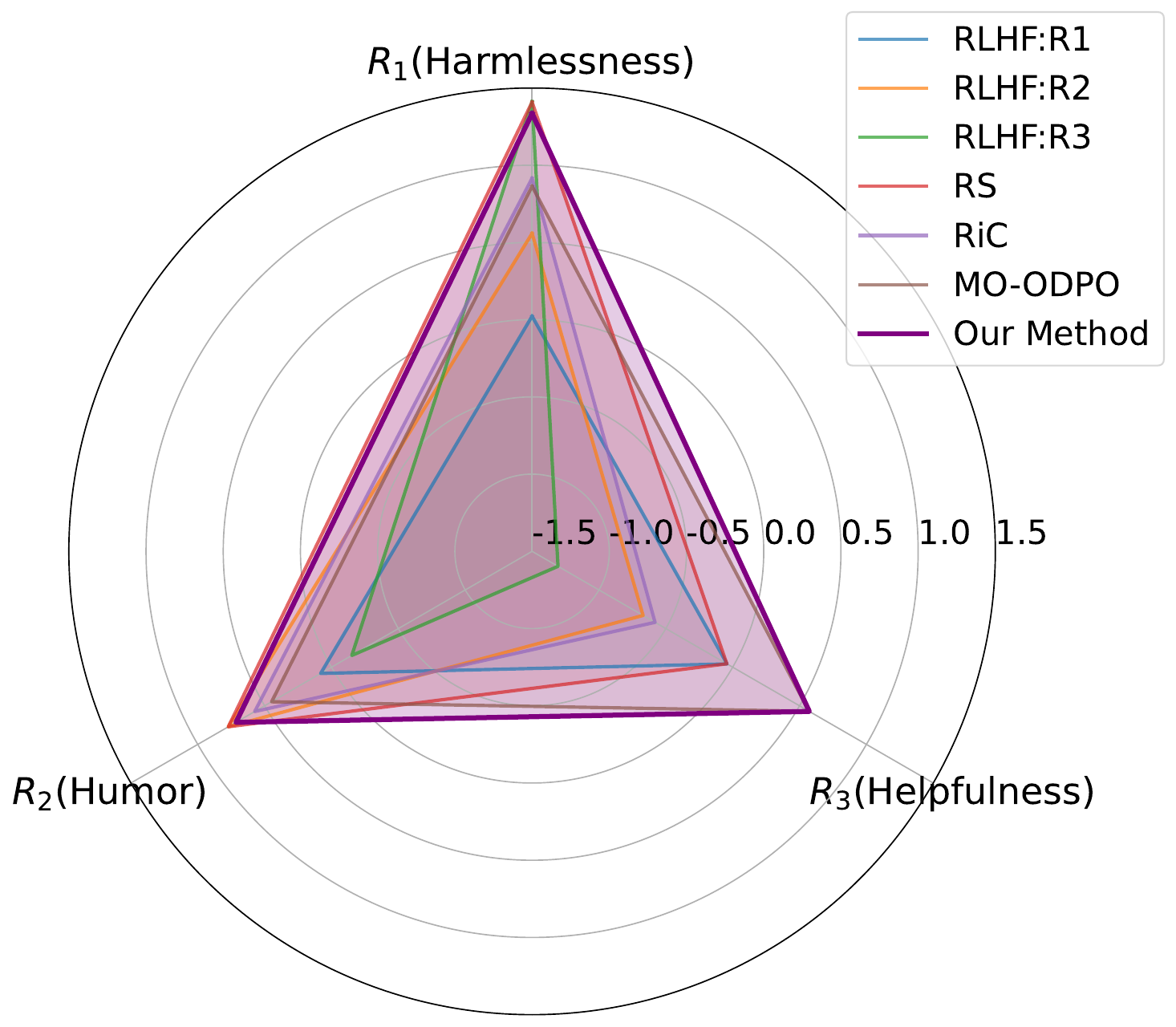}
        \caption{Optimal results of multi-objective algorithms under different preferences.}
        \label{fig:hh_radia}
    \end{subfigure}
    \caption{Results for the Helpful Assistant task with three-objective alignment using normalized harmlessness, helpfulness, and humor rewards.}
    \label{fig:hh_comparison}
\end{figure*}

\paragraph{Scaling to Three Objectives.}
To evaluate the effectiveness of our method in higher-dimensional preference spaces, we consider a Helpful Assistant task with three independent objectives: Helpfulness, Harmlessness, and Humor. As illustrated in Figure \ref{fig:hh_3d}, we dynamically adjust the preference vectors of both \mymethod\ and the baseline methods during inference. Under these varying preference conditions, our method achieves a superior Pareto frontier. In Figure~\ref{fig:hh_radia} we compare \mymethod\ with baseline methods in terms of the optimal performance achievable along each preference dimension, as well as RLHF trained on individual rewards. The results demonstrate that \mymethod\ achieves a balanced and competitive performance across all dimensions, whereas the other methods exhibit suboptimal performance in at least one preference dimension.

The quantitative results presented in Table \ref{tab:hh-3d} corroborate these observations. Among the baselines, RS performs relatively well on CRD but shows weaker performance on the other metrics. This behavior stems from its parameter-interpolation strategy: by interpolating model parameters, RS fixes model weights corresponding to specific preferences, thereby avoiding the amplification of reward variance that instruction-based methods may experience due to limited conditional-following capability. However, its linear combination of parameters constrains the model’s ability to capture nonlinear representations, making it difficult to approximate the Pareto-optimal.
Notably, as the number of aligned objectives increases, the advantages of our mutual information-driven framework become more pronounced. Compared with the currently best-performing baseline method, MO-ODPO, \mymethod\ achieves an impressive relative improvement of 87.2\% in HV and a 29.6\% gain in MIP, while maintaining highly competitive CRD scores.

\section{Related Work}
\paragraph{Multi-Objective Alignment.} Human preferences are inherently multi-dimensional and sometimes conflicting, encompassing aspects such as helpfulness, harmlessness, and humor \cite{yangrewards}, and recent studies further reveal Pareto-optimal trade-offs among these objectives \cite{agnihotri2025multi}. Effectively aligning large language models (LLMs) with such diverse preference dimensions therefore remains a critical challenge. Conventional multi-objective alignment methods \cite{zhou2024beyond,xu2025reward} typically rely on linearly combining multiple reward models and retraining the LLM for each preference configuration, which incurs substantial computational cost. To alleviate this burden, some methods train specialized models for individual preference dimensions and integrate them at inference time via parameter merging \cite{rame2023rewarded}, logit aggregation \cite{shi2024decoding}, or ensemble-based refinement \cite{li-etal-2025-self-improvement}. Nevertheless, these methods still require maintaining multiple models, leading to considerable storage and inference overhead. 
To address this limitation, alternative strategies~\cite{yangrewards,yang2024metaaligner,gupta2025robust} leverage the instruction-following capability of LLMs by incorporating preference vectors or desired preference scores into the prompt to train a single policy model. Among these, MO-ODPO~\cite{gupta2025robust} combines the optimization stability of DPO with the exploratory power of online reinforcement learning to achieve effective trade-offs across multiple objectives, but it may still generate responses that are poorly aligned with the preference vectors. Recently, UniARM~\cite{xie2026uniarm} proposes a unified autoregressive reward model that jointly models multiple preference dimensions, enabling flexible test-time alignment~\cite{zhao2024hallucinations} without requiring separate reward models for each objective. In this work, we focus on reducing uncertainty during the exploration process, ensuring that sampled responses better align with the preference vectors and thereby achieve improved multi-objective trade-offs. 

\paragraph{Multi-Objective Optimization.}

Multi-objective optimization aims to identify optimal solutions in the presence of multiple, potentially conflicting objectives. Existing methods can be broadly categorized into three types: approaches that focus on identifying a single optimal solution \cite{ye2021multi,lin2024smooth}, approaches that aim to obtain a finite set of optimal solutions \cite{lin2025few}, and approaches that approximate the entire Pareto front, which may be infinite \cite{dimitriadis2025pareto,LORPMAN}.
Of particular relevance to this study are the latter approaches, which model the complete Pareto set within a single model, enabling flexible selection of optimal solutions according to diverse user preferences without retraining. These methods have been widely applied in deep learning, including Bayesian optimization \cite{lin2022pslmobo}, reinforcement learning \cite{gupta2025robust}, and model merging \cite{chen2025pareto}. In this paper, we propose \mymethod, an online reinforcement learning framework for multi-objective alignment that reduces exploration uncertainty. Different from prior contrastive learning methods \cite{li2022simctc,su2024domain,xiong2024dual} that leverage mutual information maximization for representation learning, we employ it as a mechanism for preference-conditioned policy optimization. By maximizing the mutual information between generated responses and conditioning preference vectors, \mymethod promotes more distinguishable exploration trajectories across different preference conditions, leading to improved preference alignment while preserving the diversity of the Pareto frontier and achieving better multi-objective trade-offs.

\section{Conclusion}

In this paper, we propose \fullnamemymethod\ (\mymethod), an information-theoretic framework that unifies multi-objective exploration and alignment. \mymethod\ maximizes the joint conditional mutual information among generated responses, preference feedback, and preference vectors. By doing so, it achieves objective-level preference alignment while guiding the model toward preference-aware exploration, which reduces uncertainty in the generation process. Empirical evaluations on safety alignment and helpful assistant tasks demonstrate that MI-EPO achieves strong performance across multiple metrics, including HV, MIP, and CRD.

\begin{credits}
\subsubsection{\ackname} This work was supported by the National Natural Science Foundation of China (No. 62276015 and No. 62506024) and GW2025-09.

\subsubsection{\discintname}
 The authors have no competing interests to declare that are
relevant to the content of this article.
\end{credits}

\bibliographystyle{splncs04}
\bibliography{ref}

\appendix
 
\section{ Sources of Datasets and Models}
\label{appendix:sources}

We provide a detailed introduction to the datasets used in this paper:
\begin{itemize}
    \item PKU-SafeRLHF-10K~\cite{ji2023beavertails}, which  contains 10k samples with safety preferences. The dataset includes constraints in more than ten dimensions, such as insults, immoral, crime, emotional harm, privacy, and others. They are designed for fine-grained constraint value alignment in RLHF technology
    \item HH-RLHF~\cite{bai2022training} comprises two parts: a helpfulness dataset and a harmlessness (red-teaming) dataset, with approximately 160k annotated samples in total. The helpfulness portion is constructed by engaging crowdworkers in open-ended conversations with models and selecting the responses deemed more helpful, thereby steering the dialogue toward more beneficial directions. In contrast, the harmlessness portion is collected by prompting models to generate potentially harmful responses and selecting those deemed more harmful, thereby steering the dialogue toward harmful directions.
\end{itemize}

In Table \ref{tab:sources}, we provide the sources of datasets and models used in our experiments.

\begin{table}[h]
\centering
\scriptsize
\caption{Sources of datasets and models used in our experiments.}
\label{tab:sources}
\begin{tabular}{lcc}
\toprule
& \textbf{Safety Alignment} & \textbf{Helpful Assistant}\\
\midrule
Dataset & \href{https://huggingface.co/datasets/PKU-Alignment/PKU-SafeRLHF-10K}{PKU-SafeRLHF-10K} \cite{ji2023beavertails,ji2024pku}  & \href{https://huggingface.co/datasets/Anthropic/hh-rlhf}{HH-RLHF} \cite{bai2022training}\\
Base Models & \href{https://huggingface.co/PKU-Alignment/alpaca-7b-reproduced}{Alpaca-7B} &  \href{https://huggingface.co/Qwen/Qwen3-8B-Base}{Qwen3-8B-Base} \\
Oracle Reward Models & \href{https://huggingface.co/PKU-Alignment/beaver-7b-v1.0-reward}{Helpfulness}; \href{https://huggingface.co/PKU-Alignment/beaver-7b-v1.0-cost}{Harmlessness} &\href{https://huggingface.co/Ray2333/gpt2-large-helpful-reward_model}{Helpfulness}; \href{https://huggingface.co/Ray2333/gpt2-large-harmless-reward_model}{Harmlessness}; \href{https://huggingface.co/mohameddhiab/humor-no-humor}{Humor} \\
\bottomrule
\end{tabular}
\end{table}

\section{Implementation Details}
\label{appendix:ha_hyperparameter}
Table~\ref{tab:exp_details_text_generation} summarizes the hyper-parameter settings used in our experiments. All models are based on a Transformer architecture implemented with the TRL framework and are trained on NVIDIA Tesla A100 GPUs with 40 GB memory. We adopt bfloat16 (bf16) precision for training and apply LoRA~\cite{hu2021lora} for parameter-efficient fine-tuning, with rank $r=64$, scaling factor $\alpha=128$, and a dropout rate of 0.05. The Adam optimizer is used together with a cosine learning rate scheduler, a warm-up ratio of 0.1, and a batch size of 64. During generation and evaluation, the maximum number of inference tokens is set to 128.

The training process consists of two stages. In the supervised fine-tuning (SFT) stage, models are trained for three epochs. The initial learning rate is set to $1\times10^{-6}$ for the safety alignment setting and $2\times10^{-5}$ for the helpful assistant setting. For safety alignment, we adopt a smaller learning rate since the base model has already been fine-tuned on the corresponding dataset, and the subsequent training mainly aims to adapt the model to prompts with weights provided as additional inputs. In the online alignment stage, models are further fine-tuned for two epochs, with an initial learning rate of $1\times10^{-4}$ for safety alignment and $4\times10^{-5}$ for helpful assistant. The KL divergence coefficients, $\beta_c$ and $\beta_w$, are set to $0.1$ and $0.01$, respectively. Since the reward models for different preference dimensions have varying scales, we first compute the mean and variance of each dimension’s rewards on the offline data, and then normalize the rewards during the online stage.
For baseline reproduction, we follow the official codebase and experimental settings provided by the \cite{yangrewards}.

\begin{table}[h!]%
	\centering%
	\caption{hyper-parameter settings of the experiments.}%
	\centering
	\resizebox{0.9\textwidth}{!}{
		\begin{tabular}{cc}%
			\toprule
			\multicolumn{2}{c}{\textbf{Common Hyper-parameter Settings}}                                                                                                                                                                             \\
			\midrule

            Architecture         & Transformer, Trl                                                                                                                                               \\ 
			Hardware             & NVIDIA Tesla A100           \\
            Quantization for training & bf16 \\
            Fine-tuning strategy & LoRA \cite{hu2021lora}  \\
            LoRA $r$             & 64   \\
 			LoRA alpha           & 128  \\
			LoRA dropout                                & 0.05   \\
			Optimizer                                   & Adam   \\ 
            Learning rate scheduler                     & Cosine  \\ 
            Warm up ratio                               & 0.1      \\   
   			Batch size                                  & 64        \\
            Inference tokens for generation, and evaluation  & 128   \\

            dirichlet concentration parameters $\boldsymbol{\alpha}$ & 0.5 for Safety Alignment and 1.0 for Helpful Assistant    \\
            \midrule
			\multicolumn{2}{c}{\textbf{SFT Stage}}              \\       
			\midrule
             Finetuning epochs & 3   \\
			Initial learning rate        & $1\times10^{-6}$ for Safety Alignment and $2\times10^{-5}$ for Helpful Assistant  \\
			\midrule
			\multicolumn{2}{c}{\textbf{Online Stage}}              \\       
			\midrule
            Online finetuning epochs & $2$ for the two-objective setting and $3$ for the three-objective setting \\
			Initial learning rate        & $1\times10^{-4}$ for Safety Alignment and $4\times10^{-5}$ for Helpful Assistant \\
            $\beta_c$       & 0.1 \\ 
            $\beta_w$       & 0.01 \\ 
			\bottomrule
		\end{tabular}
	}
	\label{tab:exp_details_text_generation}
\end{table}%

\section{Details of Evaluation Metrics}

\label{appendix:eval_details}

We employ three multi-objective optimization metrics for quantitative evaluations:  hypervolume (HV)~\cite{zitzler1998multiobjective}, mean inner product (MIP)~\cite{lin2025parm} and conditional reward dispersion(CRD).
\paragraph{HV.}
Let $\mathbf{u} \in \mathbb{R}^k$ denote a solution’s objective vector, $\mathcal{A} = \{\mathbf{u}_1, \dots, \mathbf{u}_N\}$ a set of such vectors, and $\mathbf{z}$ a reference point. The hypervolume of $\mathcal{A}$ with respect to $\mathbf{z}$ is defined as
\[
\mathrm{HV}_{\mathbf{z}}(\mathcal{A}) = \Lambda \Big( \mathbf{v} \;\big|\; \exists \mathbf{u} \in \mathcal{A} : \mathbf{u} \preceq \mathbf{v} \preceq \mathbf{z} \Big),
\]
where $\Lambda(\cdot)$ denotes the Lebesgue measure.

HV quantifies the volume of objective space dominated by $\mathcal{A}$ relative to $\mathbf{z}$, capturing both convergence toward the Pareto front and diversity across objectives. A larger HV indicates better overall performance in terms of both quality and coverage.

\paragraph{MIP.} MIP is a metric used to evaluate the alignment between response and user preferences. Let $\mathbf{w}_i \in \mathbb{R}^K$ denote the user preference vector associated with the $i$-th sample, and $\mathbf{s}_i \in \mathbb{R}^K$ denote the corresponding model response reward vector. MIP is defined as
\begin{align}
\text{MIP} = \frac{1}{N} \sum_{i=1}^{N} \mathbf{w}_i^\top \mathbf{s}_i,
\end{align}
where $N$ is the total number of samples.

Intuitively, MIP measures the alignment between the model’s responses and the user’s preferences, with higher values indicating better conformity. In multi-objective alignment scenarios, each dimension of the preference vector represents a distinct preference aspect, so MIP effectively captures the model’s performance across different preference dimensions.

\paragraph{CRD.}

To evaluate the stability of the policy under a fixed preference condition, we measure the dispersion of the obtained multi-objective rewards within each condition. Specifically, let $\mathbf{s} \in \mathbb{R}^k$ denote the reward vector corresponding to a generated response. For a given preference condition $w$, we collect the evaluation vectors of all generated responses and compute the covariance matrix:
\begin{align}
    \Sigma_w = \text{Cov}(\vs \mid \vw).
\end{align}

We then define the Conditional Reward Dispersion as the determinant of the covariance matrix:
\begin{align}
    D(w) = \det(\Sigma_w).
\end{align}

The determinant of the covariance matrix, also known as the generalized variance, reflects the joint dispersion of the reward distribution. Geometrically, it corresponds to the volume (area in the two-dimensional case) of the reward distribution ellipse.

A smaller value of $D(w)$ indicates that the generated responses yield more concentrated reward outcomes under the given preference condition, suggesting that the policy exhibits more stable and consistent conditional control. Conversely, a larger value implies higher variability in reward outcomes, indicating weaker control over the target preference.

In our evaluation, we compute this quantity for each preference condition and report the average value across all conditions:
\begin{align}
    \text{CRD} = \frac{1}{|\mathcal{W}|} \sum_{w \in \mathcal{W}} \det(\Sigma_w).
\end{align}

Lower values of CRD indicate better conditional stability.

\paragraph{Evaluation Procedure} We evaluate all methods on the test dataset using preference vectors. In the two-preference-dimension experiment, we select a set of discretized and evenly spaced preference vectors on the 2D probability simplex: {[0.0, 1.0], [0.1, 0.9], [0.2, 0.8], [0.3, 0.7], [0.4, 0.6], [0.5, 0.5], [0.6, 0.4], [0.7, 0.3], [0.8, 0.2], [0.9, 0.1], [1.0, 0.0]}. 
In the three-preference-dimension experiment, we select the following representative preference vectors on the 3D probability simplex:
{[0.0, 0.0, 1.0], [0.0, 1.0, 0.0], [0.1, 0.1, 0.8], [0.1, 0.8, 0.1], [0.2, 0.2, 0.6], [0.2, 0.4, 0.4], [0.2, 0.6, 0.2], [0.33, 0.33, 0.33], [0.4, 0.2, 0.4], [0.4, 0.4, 0.2], [0.6, 0.2, 0.2], [0.8, 0.1, 0.1], [1.0, 0.0, 0.0]}, covering different regions of the trade-off space. Using these preference vectors produces a set of solutions and the corresponding discrete Pareto front for each method. In all evaluations, the rewards are normalized.

\end{document}